\documentclass{article}
\usepackage[numbers]{natbib}

\usepackage {arxiv}
\usepackage{cite}
\usepackage{floatrow}
\usepackage{amsmath,amssymb,amsfonts}
\usepackage{caption}
\usepackage{algorithm}
\usepackage{algorithmic}
\usepackage{textcomp}
\usepackage[table]{xcolor}
\usepackage{float}
\usepackage{booktabs}
\usepackage{enumitem}
\usepackage{setspace}
\usepackage[]{graphicx}
\usepackage{amsmath}
\usepackage{amssymb}
\usepackage{epsfig}
\usepackage{epstopdf}

\usepackage{subfigure}
\usepackage{caption}
\usepackage{comment}
\usepackage{multirow}

\title{\LARGE \bf
Effectiveness of Deep Image Embedding Clustering Methods on Tabular Data}

\author{Sakib Abrar, Ali Sekmen, Manar D. Samad\\
Department of Computer Science\\
 Tennessee State University\\
 Nashville, TN, USA\\
\texttt{msamad@tnstate.edu} \\
}
   
\begin{document}

%\begin{comment}

%\end{comment}

\maketitle

\begin{abstract}
 
Deep learning methods in the literature are commonly benchmarked on image data sets, which may not be suitable or effective baselines for non-image tabular data. In this paper, we take a \emph {data-centric} view to perform one of the first studies on deep embedding clustering of tabular data. Eight clustering and state-of-the-art embedding clustering methods proposed for image data sets are tested on seven tabular data sets. Our results reveal that a traditional clustering method ranks second out of eight methods and is superior to most deep embedding clustering baselines. Our observation aligns with the literature that traditional machine learning of tabular data is still a robust approach against deep learning. Therefore, state-of-the-art embedding clustering methods should consider \emph {data-centric} customization of learning architectures to become competitive baselines for tabular data. 

%traditional clustering methods are still strong baselines for tabular data, and outperforming these baselines will require us to consider data-centric customization of learning architectures. 

\end{abstract}

\keywords {tabular data, embedding clustering, image embedding, convolutional neural network, deep clustering
}

\section{INTRODUCTION}

The success of deep learning has established a strong perception that \emph{deep learning is all we need} regardless of the data problem. However, newly proposed deep learning methods are always benchmarked on several standard image data sets, including MNIST, ImageNet, and CIFAR-10~\citep{Sekmen2021}. The superiority of deep learning methods proposed for computer vision applications is often perceived as a general deep solution or baseline for other data problems. For example, an image contains homogeneous pixels with spatial regularity, which makes convolution operations in convolutional neural network (CNN) high performing and meaningful for computer vision tasks. Conversely, tabular data with heterogeneous variables can be strikingly different from image data, as shown in Table~\ref{imageVStable}. A multilayer perceptron (MLP) is a default choice for learning tabular data~\citep{Gorishniy2021} because convolution-based filtering is not intuitive on multivariate feature vectors without temporal or spatial regularity. Therefore, the contrast in data types is an important factor in designing an appropriate deep architecture.

An emerging example of deep learning application is deep embedding clustering, which has been so far proposed to obtain \emph{cluster-friendly} embedding for image data sets only~\citep{Xie2016, Guo2017, Mrabah_neunet_2020, MoradiFard2020, Boubekki2021, Yang2020, Dizaji2017}. When a novel embedding clustering method is proposed for non-image tabular data, one may expect to compare its performance with those baseline embedding clustering methods~\citep{anonymous2023gceals}. This expectation assumes that the same deep method is a fair and equally effective baseline for both image and non-image data sets. Conversely, it may be argued that the deep learning architectures proposed for image data sets may not be suitable or optimal for tabular data. To address this concern, one may take several alternative approaches. First, 2D images can be vectorized into a tabular format to form tabular data sets. However, vectorized images still contain homogeneous pixels, not multivariate feature columns. Merely converting images into pixel vectors in tables does not satisfy the characteristics of tabular data, as presented in Table~\ref{imageVStable}. Second, one may selectively use high-dimensional tabular data sets with large sample sizes to keep the problem and results compatible with image benchmarks and corresponding baseline methods. It is noteworthy that the top 100 most downloaded tabular data sets in the UCI machine learning repository have a median dimensionality of only 18, which is far below the lowest 784-dimensional pixel vectors in the MNIST data set. 
%%%%%%%%%%%%%%%%%
\begin{table*}[t]
\caption{Image versus tabular data. Median sample size and median data dimensionality are obtained across the 100 most downloaded tabular data sets from the UCI machine learning repository \citep{Dua_2019}.}
\label{imageVStable}
\begin{tabular}{lll}
\toprule
Factors      & Image data & Tabular data               \\ \midrule
Homogeneity & Homogeneous         & Heterogeneous or multivariate                  \\
Spatial Regularity & Yes & No  \\
Sample size   & Large, $>$50,000         & Small, median size $\sim{660}$        \\

Benchmark data sets     & CIFAR, MNIST         & None               \\
Dimensionality     & High, $>$1000      &   Low, median 18        \\
Best method  & Deep CNN         & Traditional machine learning           \\
Special methods         & transfer learning, image augmentation         & None          \\
Application       & Computer vision         & Data analytics             \\ 
\bottomrule
\end{tabular}
\end {table*}
%%%%%%%%%%%%%%%%

These alternative approaches are often chosen to stay within the strict realm of image benchmarks and image-based learning architectures without proposing custom methods for tabular data. Here, the \emph {model-centric} view of deep learning to achieve superior performance largely conceals the need for \emph {data-centric} requirement analysis in designing deep learning methods. A fundamental challenge for deep learning methods is to claim superiority on \emph {non-image and low-dimensional} data as these methods commonly claim on image data sets. In this context, this paper investigates the effectiveness of state-of-the-art deep-embedding clustering methods on tabular data sets. We hypothesize that these deep embedding clustering methods are not effective baselines for tabular data.   

The remainder of the manuscript is organized as follows. Section II highlights the challenges of deep learning methods on tabular data sets and state-of-the-art methods proposed for embedding clustering. Section III provides the methods and preparation of state-of-the-art embedding clustering methods for learning tabular data. Section IV highlights the important results and shows a performance comparison between traditional clustering and deep embedding clustering methods on tabular data. A summary of the results is highlighted in Section V, and this paper concludes in Section VI.

\section {Background}

\subsection {Tabular versus image data}

Table~\ref{imageVStable} summarizes the contrasts between image and tabular data. One obvious distinction is that tabular data consist of multivariate feature vectors, whereas images are a distribution of homogeneous pixels in 2D space. The pixel variable is distributed over space with spatial regularity, which makes convolution-based image filtering meaningful and effective. The variables in tabular data (e.g., age, salary, height, weight) have different scales without regularity or repetitions as pixel intensities. While images can be very high-dimensional, the number of variables in tabular data can be considerably low. The sample size of many tabular data sets is relatively much smaller than what we see in benchmark image data sets. %Therefore, the data-specific contrasts summarized in Table~\ref{imageVStable} should be considered for tailoring the baseline deep image learning methods to tabular data sets. 

\subsection {Deep learning of tabular data}

Deep learning has overtaken traditional machine learning methods because of its ability to simultaneously learn a new feature space or embedding during supervised training. This alleviates the need for suboptimal \emph{hand-engineered} features before training a machine learning model~\citep{Alam2020}. As a result, the clustering of deep embedding is known to yield significantly superior accuracy when compared to clustering on the image pixel space. Therefore, one may always expect a new deep embedding method to significantly outperform traditional machine learning.  

Conversely, it may be unknown to many computer vision researchers that traditional machine learning methods are still superior to deep learning on tabular data sets. There is mounting evidence in the literature in favor of traditional machine learning on tabular data~\citep{Kohler2019, Smith2020, Borisov2021, Kadra2021, Shwartz-Ziv2022}, including conclusions such as “deep learning is not all you need”~\citep{Shwartz-Ziv2022}, “tabular data is the last unconquered castle for deep learning”~\citep{Kadra2021}. Without reference to these studies, it may be counter-intuitive to deep learning researchers that traditional clustering methods are still strong and superior baselines for tabular data. This leads to a preconceived notion that the performance of a deep method is always expected to be significantly better than those obtained by traditional baseline methods. In practice, outperforming traditional machine learning baselines on tabular data remains a challenge for deep learning methods~\citep{Shwartz-Ziv2022, Borisov2021, Kadra2021}. This challenge remains because data-centric requirements are not usually considered when preparing learning algorithms and architectures.
%%%%%%%%%%%
\subsection {Deep Embedding Clustering}
The goal of deep embedding clustering is to learn a \emph {cluster-friendly} embedding by jointly training an unsupervised deep neural network with a clustering algorithm. Autoencoders are the most common form of unsupervised deep learning architectures used to first encoder input data (X$\in \Re^d$) into a latent space or embedding (Z$\in\Re^m$) where $d<m$. The input data are reconstructed from the embedding using a decoder module as ($\hat{X}$). The encoder and decoder involve a set of trainable parameters  $\theta\in\{ W_{\theta}, b_{\theta}\}$ and $\Phi\in\{ W_{\Phi}, b_{\Phi}\}$, respectively. The outputs of the encoder (Z) and decoder ($\hat{X}$) are obtained as follows. 
%%%%%%
\begin{eqnarray}
Z = f (\theta, X)\\
\hat {X} = g (\Phi, f (\theta, X))
\end{eqnarray}
%%%%%%%
Here, f(.) and g(.) represent sigmoid activation functions to introduce non-linearity in embedding. The learning objective of an autoencoder is to minimize the following reconstruction loss, updating $\theta$ and $\Phi$ parameters. 
%%%%%%%%
\begin{equation}
\mathcal{L}_{recon}  = \underset{\theta, \Phi}{\operatorname{argmin}} \sum^{N}_{i=1} ||  X_i - \hat{X_i} ||_2^2.
 \label{equation-LLE2}
\end{equation}
%%%%%%%%%%%
The autoencoder embedding (Z) is known to retain all information about the input data to facilitate perfect data reconstruction. Instead, an ideal embedding should emphasize or retain information useful for clustering. Accordingly, the quality of an embedding can be improved by first clustering the Z space using a clustering algorithm. The cluster assignments and centroids are used to compute an embedding distribution (Q) or pseudo labels as a learning target. A target distribution (P) is mathematically derived from the embedding distribution (Q). The overall learning objective is to update the autoencoder's trainable parameters ($\theta, \Phi$) by jointly minimizing the reconstruction loss and the divergence between P and Q distributions, as below. 
%%%%%%%
\begin {eqnarray}
\mathcal{L} &=& \mathcal{L}_{recon} + \gamma* \mathcal{L}_{cluster}  \nonumber\\ 
&=& \underset{\theta, \Phi}{\operatorname{argmin}} \sum^{N}_{i=1} ||  X_i - \hat{X_i} ||_2^2 + \gamma \sum_{i=1}^N \sum_{j=1}^K p_{ij} log \frac{p_{ij}}{q_{ij}}
\end {eqnarray}
%%%%%%%%%
Here, N is the number of samples, K denotes the number of clusters, and $\gamma$ is the trade-off parameter between the reconstruction and clustering losses.

It has been shown that the clustering accuracy on Z obtained via joint learning in Equation (4) is substantially better than that obtained by minimizing the reconstruction loss alone (Equation 3). Deep Embedding Clustering (DEC) is one of the first methods in this area~\citep{Xie2016}. The DEC method first pretrains a deep autoencoder by minimizing the data reconstruction loss only. Excluding the decoder, the pre-trained encoder part  is then fine-tuned by minimizing the Kullback-Leibler (KL) divergence between a t-distributed cluster distribution (Q) on the embedding and a target distribution (P). The DEC method is further improved by proposing an improved DEC (IDEC) framework~\citep{Guo2017}. In IDEC, the autoencoder reconstruction loss and the KL divergence loss are jointly minimized to update the weights of a deep autoencoder for producing a cluster-friendly embedding (Equation 4). Both DEC and IDEC methods perform k-means clustering on embedding to obtain the t-distributed cluster distribution (Q). A joint embedding and cluster learning (JECL) is proposed for multimodal representation learning of text-image data pairs~\citep{Yang2020} using t-distribution assumption, k-means clustering, and KL divergence loss between the embedding and target distributions.  The Deep Clustering via Joint Convolutional Autoencoder (DEPICT) method is proposed for learning image embedding via a de-noising convolutional autoencoder~\citep{Dizaji2017}. Following this research trend, new embedding clustering methods have been proposed in recent years ~\citep{Mrabah_neunet_2020, MoradiFard2020, Zhang2019, Boubekki2021}, which are all evaluated on benchmark image data sets.

\begin{table}[]
%\scalebox{0.85}{
\caption{Summary of tabular data sets used for comparing clustering and embedding clustering performance in this study.}
\begin{tabular}{lccc}
\toprule
Data set      & Sample size & Dimensions & Classes %& Domain  
\\ \midrule
Breast Cancer & 569         & 30         & 2  \\           % & Diagnostic        \\
Dermatology   & 358         & 34         & 6   \\          % & Histopathological \\
E. coli       & 336         & 7          & 8    \\%          & Protein cell      \\
Malware       & 4465        & 241        & 2      \\%        & Android malware      \\
Mice data  & 552         & 78         & 8          \\%    & Protein expression  \\
Olive         & 572         & 10         & 3         \\%     & Food \& beverage  \\
Vehicle       & 846         & 18         & 4        \\%      & Silhouette features    \\ 
\bottomrule
\end{tabular}
\label{summary}
%}
\end{table}
\section {Methods}

Because current deep embedding clustering methods are evaluated only on image data sets, several image-specific operations are often involved, including image augmentation, Sobel filtering, transfer learning, image corrupting, and denoising. These operations are not trivial on tabular data sets. CNN-based architectures with 2D image filtering are proposed for embedding clustering in addition to several MLP-based architectures. The 2D image filters in CNN are at least required to be replaced by 1D kernels for tabular data. Therefore, deep baseline methods proposed for image learning require modifications for learning tabular data. 

\subsection {Baseline embedding clustering methods}

We compare the clustering performance of six state-of-the-art deep embedding clustering methods with traditional clustering (K-means, Gaussian mixture model) of tabular data. The deep embedding clustering models are trained for a varying number of epochs, which we set to 1000 epochs to fairly compare the methods. We make the least possible changes to the state-of-the-art embedding clustering methods for producing embeddings on tabular data sets as follows.

\subsubsection{DEC method}

The autoencoder architecture in the DEC~\citep{Xie2016} method uses a fully-connected autoencoder with three hidden layers. The encoder and decoder sections are set to d–500–500–2000–10-2000-500-500-d, inspired by~\citep{van2009learning}. Here, d is the input data dimension, 500 or 2000 denotes the number of neurons at a given layer, and the embedding dimensionality is 10. We have disabled the greedy layer-wise pretraining done for 50000 iterations and the 20\% dropout in layers to make a fair comparison with other methods.

\subsubsection{IDEC method}
The Improved DEC (IDEC)~\citep{Guo2017} and DEC methods share the same architecture. While the DEC method optimizes the reconstruction loss of the autoencoder and the clustering loss separately, IDEC optimizes both losses jointly for the first time, as shown in Equation 4.

\subsubsection{DKM method}

The deep k-means (DKM) method uses an autoencoder architecture similar to the DEC and IDEC methods ~\citep{MoradiFard2020}. Instead of a fixed dimensional embedding, they use a k-dimensional latent space (d–500–500–2000–k-2000-500-500-d) where k is the number of target clusters.  The DKM method does not use cluster distributions or KL divergence loss as the DEC method. Instead, the authors propose a clustering loss that minimizes the distance between the embedding and cluster representatives. They compare the DKM method only against MLP-based models to avoid architectural bias.

\subsubsection{AE-CM method} We use the embedding clustering method proposed by Boubekki et al., which has a clustering module (CM) integrated into a deep autoencoder (AE) for joint learning of the embedding~\citep{Boubekki2021}. The AE-CM method uses a fully connected autoencoder (d-500-500-2000-p-2000-500-500-d) with leaky rectified linear unit (ReLU) activations where p is the dimension of the embedding. We use the default setting of this method where p is set to k, the number of target clusters. The three learning hyperparameters tuned are denoted as $\alpha, \beta, \lambda$. The method is implemented using the Keras deep learning package. 

\subsubsection{DynAE method} The DynAE method is also implemented using the Keras deep learning package~\citep{Mrabah2020}. The deep autoencoder architecture is similar to the DEC/IDEC method d-500-500-2000-10-2000-500-500-d. However, the objective function is regularized by image augmentation and an adversarially constrained interpolation step. The image augmentation (shifting and rotation) is disabled for producing tabular data embeddings. Their default training and pretraining steps are performed for 130000 and 100000 epochs. Instead, we decide to skip the pertaining step for the DynAE and other methods to ensure a fair comparison.

%All layers except for the bottleneck layer and the last layer, rely on ReLu activation functions.

\subsubsection{DEPICT method}The DEPICT method uses a convolutional autoencoder with 2D image filters and is implemented using the Theano deep learning package~\citep{Dizaji2017}. We have replaced the 2D filters with 1D kernels for learning embedding from tabular data vectors. Following three convolutional layers, the autoencoder has a fully-connected network with a d-50-50-10-50-50-d architecture. Because the Theano library support and upgrade have been discontinued for the last two years, we have to update the code for each neural network layer and the loss function to accept float64 input instead of float32.

\subsection {Data sets}
A summary of the seven tabular data sets is provided in table~\ref{summary}, which are representative of seven application domains. These data sets include examples of different sample sizes, data dimensionality, and class distributions, which may affect the performance of embedding clustering methods. Except for the malware data set, all data sets have less than 1000 samples. The number of variables or data dimensionality ranges between seven and 241, which is far below the pixel dimension of any image data set.  

\begin{figure}[t]
\centering
%\vspace{-10pt}
 \includegraphics[trim=0cm 0cm 0cm 0cm, width=0.7\textwidth] {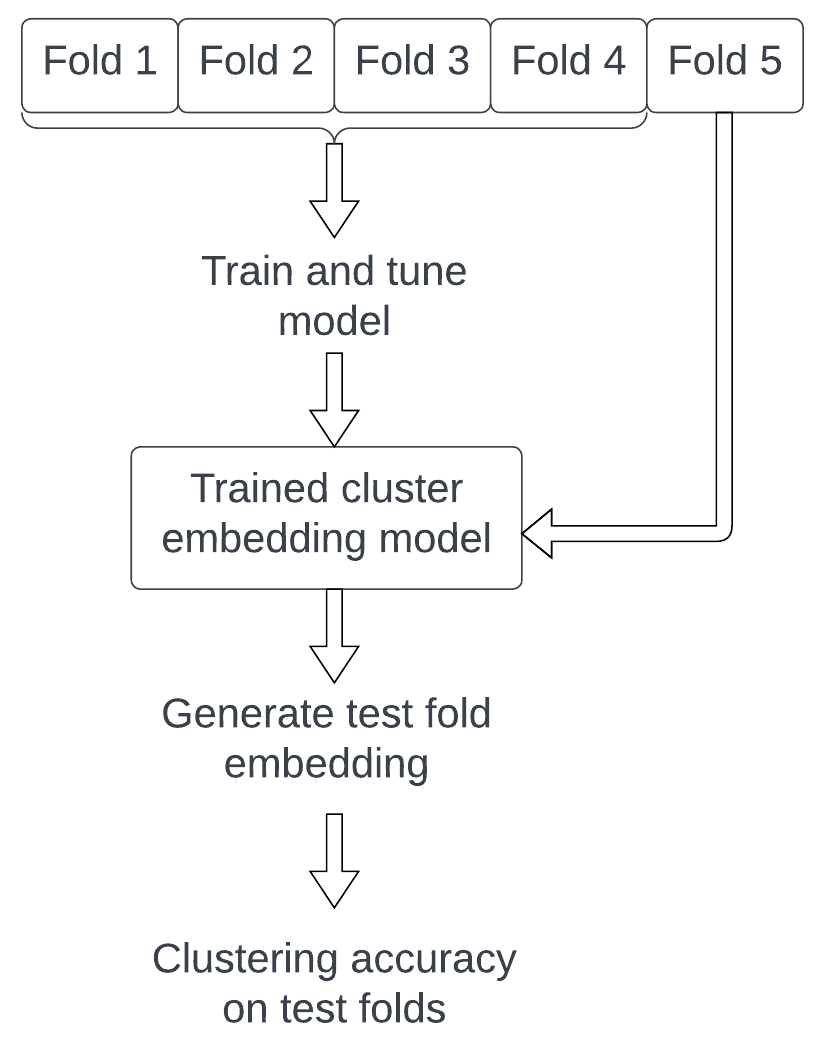}
\caption {Five-fold validation and evaluation of deep embedding clustering methods for reproducibility.}
\label{eval}
\end{figure}

\begin{table*}[]
\begin{tabular}{@{}lcccccccc@{}}
\toprule
Data set       & GMM    & K-means & DEC         & IDEC        & AE-CM       & DynAE       & DEPICT      & DKM         \\ \midrule
Breast cancer & 89.8 (4.6)  & 90.2 (4.3)   & 68.3 (5.1)  & 83.6 (14.0) & 77.7 (14.4) & 92.1 (3.8)  & 91.2 (0.5)  & 64.2 (3.9)  \\ \midrule
Dermatology   & 76.8 (8.8)  & 76.2 (9.2)   & 50.0 (5.6)  & 78.8 (10.3) & 55.5 (4.2)  & 77.4 (2.3)  & 62.3 (5.8)  & 23.2 (0.5)  \\ \midrule
Ecoli         & 29.4 (4.5)  & 29.2 (3.2)   & 29.2 (3.1)  & 32.4 (2.5)  & 30.7 (2.0)  & 31.5 (3.2)  & 35.7 (4.4)  & 35.4 (2.9)  \\ \midrule
Malware       & 79.1 (1.7)  & 79.1 (1.7)   & 78.9 (1.3)  & 85.8 (3.2)  & 73.3 (7.5)  & 73.1 (4.7)  & 56.3 (2.1)  & 48.7 (11.3) \\ \midrule
Mice data     & 40.8 (1.7)  & 40.2 (5.7)   & 36.2 (3.9)  & 40.6 (1.7)  & 36.0 (3.4)    & 42.2 (3.5)  & 25.3 (3.4)  & 17.2 (2.7)  \\ \midrule
Olive         & 67.2 (11.5) & 71.4 (10.8)  & 68.0 (10.0) & 77.4 (3.9)  & 53.7 (8.8)  & 47.9 (4.0)  & 55.3 (7.1)  & 56.5 (0.0)  \\ \midrule
Vehicle       & 39.4 (2.9)  & 37.2 (1.7)   & 38.3 (4.3)  & 42.4 (3.3)  & 37.5 (2.4)  & 37.2 (1.2)  & 37.2 (0.6)  & 31.1 (5.6)  \\ \midrule
%Average       & 60.4 (21.8) & 60.5 (22.5)  & 52.7 (17.74 & 63.0 (21.6) & 52.1 (17.1) & 57.3 (21.2) & 51.9 (20.2) & 39.5 (16.2) \\ \bottomrule
\end{tabular}
\caption{Five-fold average clustering accuracy of traditional clustering algorithms (Gaussian mixture model: GMM, k-means) and six deep embedding clustering methods on tabular data sets.}
\label{acc_table}
\end{table*}

\begin{table*}[]
\begin{tabular}{@{}lcccccccc@{}}
\toprule
Data set       & GMM                     & K-means                 & DEC                           & IDEC                          & AE-CM                         & DynAE                         & DEPICT                        & DKM                           \\ \midrule
Breast Cancer & 4                             & 3                             & 7                             & 5                             & 6                             & 1                             & 2                             & 8                             \\ %\midrule
Dermatology   & 3                             & 4                             & 7                             & 1                             & 6                             & 2                             & 5                             & 8                             \\ %\midrule
Ecoli         & 6                             & 8                             & 7                             & 3                             & 5                             & 4                             & 1                             & 2                             \\ %\midrule
Malware       & 2                             & 3                             & 4                             & 1                             & 5                             & 6                             & 7                             & 8                             \\ %\midrule
Mice protein     & 2                             & 4                             & 5                             & 3                             & 6                             & 1                             & 7                             & 8                             \\ %\midrule
Olive         & 4                             & 2                             & 3                             & 1                             & 7                             & 8                             & 6                             & 5                             \\ %\midrule
Vehicle       & 2                             & 7                             & 3                             & 1                             & 4                             & 6                             & 5                             & 8                             \\ \midrule
Average       & 3.3 (1.4) & 4.4 (2.1) & 5.1 (1.7) & 2.2 (1.5) & 5.6 (0.9) & 4.0 (2.6) & 4.7 (2.2) & 6.7 (2.2) \\ 
Overall rank & 2 & 4& 6 & 1& 7 &3& 5& 8 \\
\bottomrule
\end{tabular}
\caption{Tabular data set-specific and overall rank ordering of clustering and deep embedding clustering methods.}
\label{rank_table}
\end{table*}

\subsection {Model evaluation}

The quality of tabular data embedding is evaluated using the clustering accuracy metric, as shown below.

\begin{equation}
ACC  = \underset{m}{\operatorname{max}} \frac{\sum^{N}_{i=1} 1 \{y_{true}(i) = m(y_{pred}(i))\}}{N}
 \label{equation-LLE2}
\end{equation}

Here, $y_{true}$(i) is the ground truth label for the i-th sample. $y_{true}$(i) is the predicted label following clustering. m() finds the best label mapping between the cluster and the ground truth labels. The mapping can be obtained by the Hungarian algorithm~\citep{Kuhn1955}. The accuracy score is multiplied by 100 to represent scores in percentages. In all experiments, the number of clusters equals the number of known classes for individual data sets (Table~\ref{summary}). Although embedding clustering methods are unsupervised, the cluster accuracy metric uses ground truth labels to determine the accuracy of cluster assignments. We use separate training and test data folds to ensure reproducibility and tune the model hyperparameters (e.g., $\gamma$ in Equation 4) on the cluster accuracy metric. A five-fold validation scheme is used where four folds are used for training and tuning the model hyperparameters, as shown in Figure~\ref{eval}. The weakly supervised trained model with the best hyperparameter setting, obtained using four data folds, is then used to generate the embedding on the test data fold. The clustering accuracies on the left-out test folds are averaged to report and compare the final performance of an embedding clustering method.

\section {Results}

Table~\ref{acc_table} compares the clustering accuracy of eight clustering and deep embedding clustering methods. The traditional clustering Gaussian mixture model (GMM) ranks the second best for the malware, mice, and vehicle data sets. The k-means clustering of tabular data ranks the second best for the olive data set. GMM and k-means clustering rank the third best on the dermatology and breast cancer data sets, respectively. The GMM clustering algorithm ranks the second best among eight methods with an average rank of 3.3 (1.4) across seven tabular data sets, as shown in Table~\ref{rank_table}.

The clustering accuracy of the first embedding clustering method (DEC) is substantially worse than other methods for breast cancer (68\% versus 90\%) and dermatology (50\% versus 76\%) data sets. Although the DEC method ranks third for the olive and vehicle data sets, its average rank is 5.1(1.7), and the overall rank is six out of eight methods. The recent deep embedding clustering method (AE-CM) yields substantially worse performance on most tabular data sets, including the olive (54\% versus k-means: 71\%) and malware (73\% versus k-means: 79\%) data sets. The AE-CM method earns an average rank of 5.6 (0.9) and an overall rank of seven out of eight methods. The dynAE method ranks the third best among the eight methods with an average rank of 4.0 (2.6). dynAE is the best method for the breast cancer and mice data sets, and it ranks the second best for the dermatology data set. However, its performance improvement appears marginally better than the k-means clustering for the breast cancer (92.1\% versus K-means 90.2\%) and mice (42.2\% versus k-means: 40.2\%) data sets.

% \usepackage{booktabs}
\begin{table}[]
\begin{tabular}{@{}ccccccccc@{}}

\end{tabular}
\end{table}

The CNN-based embedding clustering method DEPICT yields the best clustering accuracy on the Ecoli data set and the second best on the breast cancer data set. The improvement on the Ecoli data set is notable (35.7\% versus k-means: 29.2\%). However, the DEPICT method yields substantially worse performance on the dermatology (62.3\% versus k-means 76.2\%), malware (56.3\% versus k-means 79.1\%), mice (25.3\% versus k-means 40.2\%), olive (55.3\% versus 71.4\%) data sets. The average rank of the DEPICT method is 4.7 (2.2), and the overall rank is five out of eight methods. The DKM method ranks the worst among the eight deep embedding clustering methods, with an average rank of 6.7 (2.2). However, it achieves the second-best performance on the Ecoli data set on par with the DEPICT method. 

The IDEC method, proposed in 2017, ranks as the best embedding clustering method for tabular data. It ranks the best for four data sets: dermatology, malware, olive, and vehicle. The average rank of the IDEC method is 2.2 (1.5) compared to the second-best method (GMM clustering with an average rank of 3.3). Although IDEC produces substantially worse results on the breast cancer data set (83.6\% versus k-means 90.2\%), it shows some decent improvements on the malware (85.8\% versus k-means 79.1\%), olive (77.4\% versus k-means 71.4\%), and vehicle (42.4\% versus k-means 37.2\%) data sets.

% superior method on image does not 

\section {Discussion of results}

This paper is one of the first to investigate the performance of state-of-the-art deep embedding clustering methods on tabular data sets. Current deep embedding clustering methods reveal several new observations on tabular data sets. First, recently proposed or state-of-the-art embedding clustering methods are not among the best for tabular data. The best method (IDEC) is proposed in a 2017 research paper. That is, the improvement of deep embedding clustering over the years has happened by targeting benchmark image data sets only. Second, no single method is the best choice for all seven tabular data sets. Conversely, the deep learning literature often shows that the proposed method is superior to all baseline methods on all image data sets. This finding suggests that the performance of embedding clustering methods on tabular data may depend on the data domain and dimensionality. Third, traditional clustering of tabular data sets (GMM, k-means) is still a competitive baseline compared to their deep embedding clustering counterparts. While clustering on image pixel space is ineffective, the clustering of tabular data yields the second and fourth-best accuracy numbers among the eight methods we compare in this paper. Despite the promising performance of several deep embedding clustering methods against traditional clustering, the improvement is often marginal. Fourth, MLP-based architectures (IDEC, DynAE) are superior to the CNN-based method (DEPICT) on tabular data sets. Similar CNN-based architectures proposed for image learning may be replaced by MLP-based architectures for tabular data sets. Fifth, clustering objective functions that use cluster distributions (IDEC) are superior on tabular data sets to using cluster centroids (DKM) as pseudo labels. The performance of the two worst methods (AE-CM and DKM) may be attributed to the dimension of the embedding, which is set to the number of clusters. 
  
It is important to note that the comparison of deep embedding clustering algorithms can be biased by architectural selections (convolutional autoencoder versus denoising autoencoder versus stacked autoencoder, deep versus shallow autoencoder) of individual methods. Furthermore, other secondary methods, including model pertaining, dropout learning, image processing or augmentation, can play an important role in improving the clustering accuracy beyond the primary contribution to the learning algorithm or objective function. Therefore, making a fair comparison among the baseline algorithm independent of those secondary steps or architectural bias can be challenging, as we perform in this paper. Our results reveal that state-of-the-art deep embedding clustering methods may not be fair and effective baselines for comparing similar methods proposed for tabular data sets.

\section {Conclusions}
This paper compares six state-of-the-art deep embedding clustering methods with traditional clustering on tabular data sets. Our results reveal that deep methods benchmarked on image data sets are often not optimal for learning tabular data. In contrast, traditional clustering on tabular data sets is still a superior baseline to most deep embedding clustering methods. Therefore, data-centric requirement analysis should be considered in developing a deep learning method superior to those traditional baselines. 

\section*{Acknowledgements}
This research is supported by a Department of Defense (DoD) Grant W911NF-20-1-0284.

\bibliographystyle{unsrt}
\bibliography{mybib}

\end{document}